\documentclass[11pt]{article}

\usepackage[final]{acl}

\usepackage{times}
\usepackage{latexsym}
\usepackage{booktabs}
\usepackage{amsmath}

\usepackage[T1]{fontenc}
\usepackage[utf8]{inputenc}

\usepackage{microtype}

\usepackage{inconsolata}

\usepackage{graphicx}

\title{Accurate Legal Reasoning at Scale: Neuro-Symbolic Offloading and Structural Auditability for Robust Legal Adjudication}

\author{Stanisław Sójka \and Witold Kowalczyk \\
  Delos AI Inc. \\
  \texttt{\{stan, witold\}@delosintelligence.com} \\}

\begin{document}
\maketitle

\begin{abstract}

Legal texts often contain computational legal clauses---provisions whose understanding requires complex logic. While frontier Large Reasoning Models (LRMs) can describe such clauses, building production-ready systems is limited by reasoning errors and the high cost of inference. We propose Amortized Intelligence, a neuro-symbolic approach where we use an LLM once to translate a legal text into Deterministic Autonomous Contract Language (DACL): a typed graph intermediate representation. Adjudication then relies on deterministic graph executions with a visually auditable trace. In comparison against runtime LRM baselines (including GPT-5.2 and Gemini 3 Pro), our DACL-based Agent achieves near-perfect consistency and mitigates the "reasoning cliff" observed in probabilistic models. The system reduces compute costs by over $90\%$ in high-volume workflows while satisfying the strict auditability requirements of legal adjudication.

\end{abstract}

\section{Introduction}

Legal texts---laws, regulations, and commercial contracts---are composed of complex clauses that define rights and obligations. Often, these provisions exceed the semantic complexity of the real-world activities they govern. Correctly interpreting them requires a synthesis of mathematical reasoning (e.g., tax formulas), logical condition evaluation (e.g., eligibility thresholds), and contextual language interpretation. We refer to such provisions as \emph{Computational Legal Clauses}. They are widespread across high-stakes domains including civil procedure, insurance, taxation, and financial regulation.

The landscape of legal AI has expanded rapidly from discriminative tasks like judgment prediction \cite{Aletras2016PredictingJD, katz2017general, chalkidis-etal-2019-neural, chalkidis-etal-2022-lexglue, t-y-s-s-etal-2023-zero} and automated contract review \cite{hendrycks2021cuad, leivaditi2020benchmark} to generative applications powered by Large Language Models (LLMs)\cite{hou-etal-2025-clerc}. 

However, while specialized models demonstrate impressive semantic fluency \cite{Huang-LLaMA, Cui2023ChatlawAM}, they often lack the structural rigor required for statutory reasoning \citep{blair2023statutory}. 
Recent benchmarking \cite{Guha2023,fan2026lexambenchmarkinglegalreasoning} reveals that even frontier models struggle with structural rule application, while specialized evaluations like LexNum \cite{zhang-etal-2025-legal} highlight critical failures in 'legal numeracy'—the precise intersection of procedural logic and arithmetic \citep{chen2023program,gao2023pal}.

In production-ready adjudication, this probabilistic nature creates two prohibitive barriers: reliability and economics. 
First, LLMs can fail silently on arithmetic---hallucinating pricing coefficients or dropping nested conditions, a phenomenon often linked to the unfaithfulness of chain-of-thought reasoning \citep{Turpin2023}. In high-stakes contracts, a calculation cannot yield non-deterministic outputs across identical inputs. Second, relying on Chain-of-Thought (CoT) inference for every recurring transaction (e.g., processing thousands of monthly invoices) forces a linear cost scaling \citep{hsieh-etal-2023-distilling, Chen2023FrugalGPTHT}. This makes autonomous adjudication commercially unattractive for high-volume settings.

\begin{figure*}[t!]
    \centering
    \includegraphics[width=0.80\textwidth]{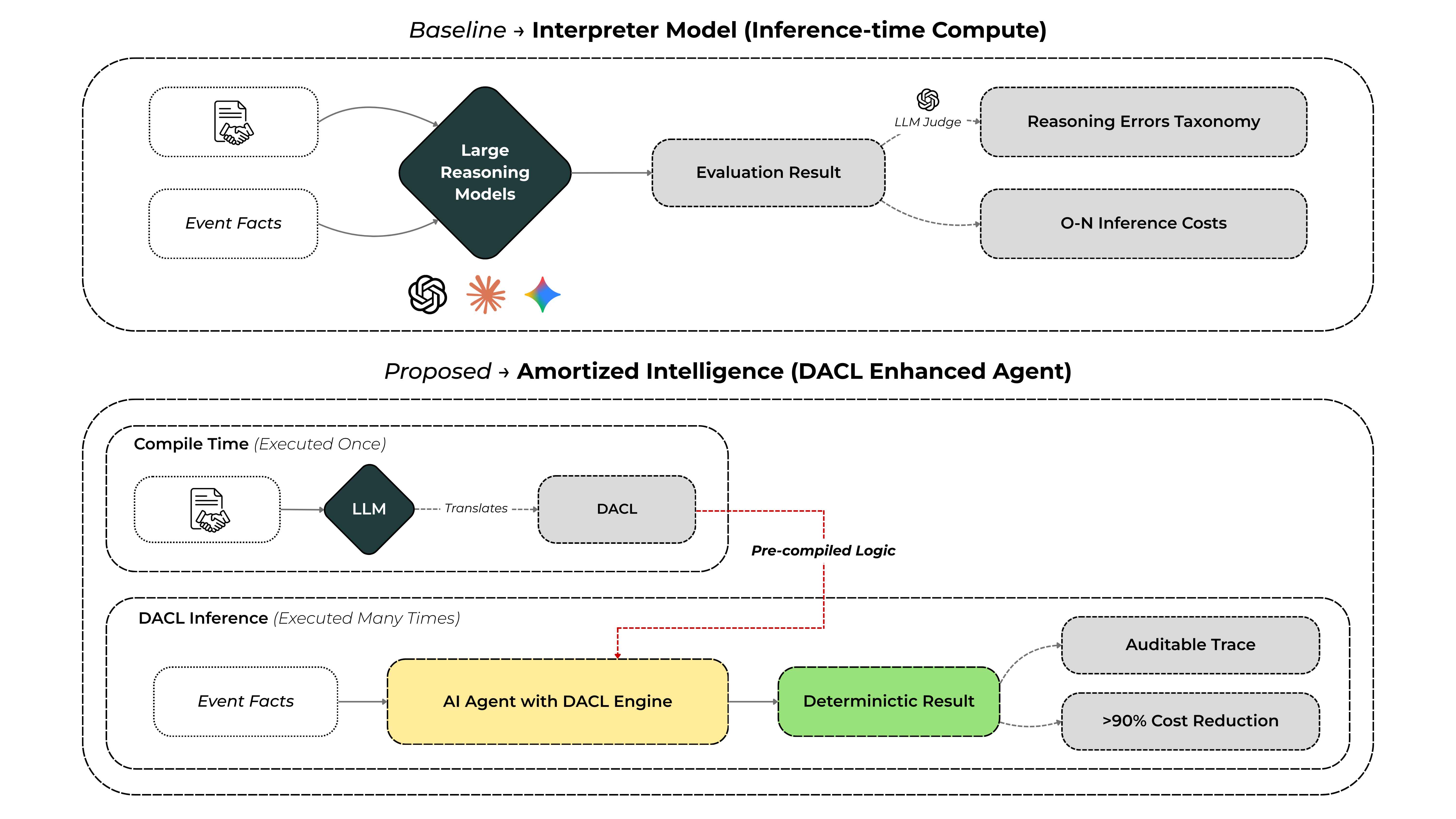}
    \caption{\textbf{Architectural Comparison: Baseline vs. Amortized Intelligence.} 
    The baseline (top) relies on expensive, probabilistic \textit{inference-time compute} for every transaction, leading to linear cost scaling and potential hallucinations. 
    Our proposed approach (bottom) shifts reasoning to a one-time \textit{compile-time} step, translating the contract into a DACL graph. This enables the runtime agent to execute deterministic logic with an auditable trace.}
    \label{fig:architecture_overview}
\end{figure*}

Foundational rigor in computational law has traditionally relied on logic programming, where Prolog \cite{KornerProlog2022} and its extensions—such as ProbLog \cite{DeRaedtProbLog2007}, EPILOG \cite{PortoEpilog1984}, and Blawx \cite{MorrisBlawx2023}—enable defeasible reasoning \cite{AntoniouBikakisDRProlog2007} but suffer from a manual knowledge acquisition bottleneck. Recent neuro-symbolic research addresses this by employing LLMs to bridge unstructured text and structured logic. Systems like SOLAR \cite{SadowskiSOLAR2025} and ProSLM \cite{Vakharia2024} decompose reasoning into extraction and application stages, while \citet{Kant2024, kant2025robust} demonstrate effective policy-to-Prolog translation, with the latter
introducing a guided encoding framework for improved adjudication
accuracy. In commercial contexts, focus has shifted to automating semantic graph construction via reinforcement learning \cite{DechtiarGRPO2025} and ensuring structural validity through uncertainty quantification \cite{DechtiarEval2025} and SMT-verified logical consistency \cite{SadowskiPrompting2025}.

While these systems advance the state of the art, they often remain fragmented—focusing on static graph extraction or theoretical simulations rather than the high-volume transaction processing required by industry. Many approaches simplify the legal problem space, omitting computationally complex components like real date arithmetic or full rule logic, and fail to address practical constraints on inference latency and cost.

To resolve these trade-offs, we introduce \emph{Amortized Intelligence}: a neuro-symbolic architecture that shifts the paradigm from a \emph{run-time interpreter} to a \emph{logic compiler}. As illustrated in Figure \ref{fig:architecture_overview}, we use an LLM only once to translate a contract into \textsc{DACL}, a deterministic intermediate representation. Subsequent adjudications reduce to efficient graph execution. Unlike general-purpose logic programming (e.g., Prolog) optimized for theorem proving, \textsc{DACL} is engineered specifically for commercial logic, employing a typed abstract syntax tree with primitives for pricing formulas, dates, and procedural clauses.
Compilation also makes interpretive choices on quantifiable
provisions explicit and stable across transactions, so that
identical inputs produce identical outputs by construction. We
evaluate this design across four business domains, four clause
primitives, and four LRM configurations.

The contributions of this paper are as follows:
\begin{itemize}
    \item \textbf{DACL Architecture:} We introduce a graph-based neuro-symbolic intermediate representation optimized for commercial logic. \textsc{DACL} prioritizes legal primitives and explicit variable-dependency tracking over general theorem proving.
    
    \item \textbf{Amortized Intelligence:} We validate our approach on data sampled from industrial Enterprise Resource Planning/accounting systems and demonstrate the advantage of computational off-loading. By handling heavy legal interpretation once and subsequently running agentic inference on the resulting \textsc{DACL} graph, we achieve $99.5\%$ accuracy on all tasks and $>95\%$ cost reduction for high-volume tasks over the closest performing LRM setup.
    
    \item \textbf{Taxonomy of Failure \& Structural Auditability:} We analyze the limitations of frontier Large Reasoning Models in legal arithmetic, isolating error modes. \textsc{DACL} provides a deterministic, traceable execution path for every decision, satisfying explainability requirements essential for legal compliance.

    \item \textbf{Production Deployment:} The DACL Agent has been deployed in a live production environment for over 12 months, where it autonomously executes billing and accounts payable workflows across 150+ commercial agreements and approximately 1,000 monthly billing events. The system operates continuously under real-world constraints and enterprise compliance requirements, replacing manual billing processes and delivering sustained, measurable cost reductions.
\end{itemize}

\section{Method}
\label{sec:method}

We evaluate two distinct architectures for automating industrial contract execution: a direct \textit{Probabilistic Interpreter} (baseline) and a \textit{Neuro-Symbolic Orchestrator} (DACL). The baseline represents the prevailing paradigm where a general-purpose frontier model reasons over the full contract text at runtime. In contrast, our proposed approach decouples reasoning from execution: the contract is compiled into a deterministic intermediate representation (DACL), which is then executed by a lightweight agent equipped with a symbolic engine.

\subsection{Baseline: Probabilistic Interpretation}
\label{subsec:baseline}

The baseline system operates as a stateless interpreter. For every input event $e_i$, the system constructs a prompt containing the full natural language text of the relevant contract $C$, the specific runtime facts $F_i$ (e.g., shipment weight, transaction date), and the query $Q$. This approach incurs $O(N)$ inference costs, as the model must re-read and re-interpret the contract for every transaction.

To ensure a rigorous comparison against state-of-the-art capabilities, we benchmarked against three frontier model families (GPT-5.2, Claude 4.5 Sonnet, and Gemini 3 Pro), explicitly enabling their respective extended reasoning modes where available. Full configuration details are provided in Appendix \ref{sec:models}.

\subsection{Symbolic Execution Engine (DACL)}
\label{subsec:dacl_engine}

The core of our proposed architecture is the DACL, a domain-specific intermediate representation that models contract logic as a directed acyclic graph (DAG). Unlike probabilistic LLM-based approaches that suffer from semantic drift across multi-step calculations, DACL guarantees \emph{referential transparency}---the same input always produces the same output---enabling formal verification of contract computations. The translation of natural language contracts into this graph structure follows a rigorous five-stage pipeline involving agentic segmentation, classification, and verified generation (see Appendix~\ref{sec:dacl_pipeline} for full implementation details).

\subsubsection{Clause Primitive Hierarchy}
DACL utilizes a strongly-typed variable schema to ensure referential transparency. It supports four recursive clause primitives, each targeting a distinct pattern in contract logic. Detailed formal definitions and validation rules for these primitives are provided in Appendix \ref{sec:appendix_dacl}, together with the engine's specifications for temporal clause versioning, structural audit trail generation, and programmatic error recovery.

\begin{enumerate}
    \item \textbf{Procedure}: Models sequential workflows where step $t$ consumes outputs from $t-1$. Procedures support \emph{hybrid clause patterns} with early termination: when a conditional step resolves to a terminal value, subsequent steps are skipped, optimizing execution time for complex fallback logic.
    
    \item \textbf{Logical Clause}: Implements first-order propositional logic over contract variables. The engine evaluates conditions in declaration order, returning on the first match. This strictly enforces priority-based eligibility rules.
    
    \item \textbf{Range Clause}: Maps continuous variables to discrete brackets with explicit boundary semantics. By enforcing strict, non-overlapping interval validation at load time, this primitive eliminates the ``off-by-one'' failures.
    
    \item \textbf{Pricing Formula}: Executes arithmetic expressions with configurable decimal precision. Sub-formulas are evaluated sequentially in a sandboxed environment that blocks arbitrary code execution while supporting standard mathematical operations via variable interpolation.
\end{enumerate}

\paragraph{Amendments and temporal versioning.}
DACL supports clause-level recompilation: when a contract is amended,
only the affected clauses are re-translated. Each clause carries
\texttt{validity\_start\_date} and \texttt{validity\_end\_date}
annotations (Appendix \ref{sec: validity}), and the engine selects the
version in force on the event date. This prevents retroactive rating
errors without requiring full contract regeneration.

\subsection{Neuro-Symbolic Agent}
\label{subsec:dacl_agent}

The runtime orchestrator is an autonomous agent built using the \textbf{OpenAI Agents SDK} \cite{openai2025agents}, following \citep{yao2023reac, schick2023} to interleave reasoning traces with symbolic tool execution. Unlike the baseline, which relies on the LLM for arithmetic and logic, this agent uses the LLM solely for semantic routing and result synthesis.

To demonstrate the efficiency of this architecture, the agent utilizes \texttt{gpt-5-mini} \cite{openai2025mini}, a lightweight model optimized for low-latency tool use. The agent is provided with exactly one tool:

\begin{quote}
\texttt{evaluate\_clauses\_tool($K$, $F_i$)}: Invokes the symbolic engine to execute the logic blocks identified by $K$ against the current event data $F_i$.
\end{quote}

Formally, for a natural language query $Q$ and runtime facts $F_i$, the agent approximates the adjudication function $\mathcal{A}(Q, F_i)$ via the following three-stage pipeline:

\begin{enumerate}
    \item \textbf{Semantic Mapping}: Leveraging techniques from constrained semantic parsing \citep{shin-etal-2021, Geng2023GrammarConstrainedDF}, the agent first maps the unstructured query and context to a precise subset of contract clause identifiers $K \subset \mathcal{C}_{ids}$. This step isolates the relevant logic without executing it:
    \begin{equation}
        K = \mathcal{M}_{\theta_{small}}(Q, F_i)
    \end{equation}
    
    \item \textbf{Symbolic Delegation}: The agent invokes the tool to trigger the deterministic engine $\Phi_{DACL}$. This step executes the logic within the sandbox, returning both a computed value $v$ and a structural audit trace $\tau$:
    \begin{equation}
        (v, \tau) = \Phi_{DACL}(K, F_i)
    \end{equation}
    Critically, $\Phi_{DACL}$ is non-probabilistic ($P(v|K, F_i) = 1$), ensuring mathematical precision and strict adherence to contract boundaries.
    
    \item \textbf{Synthesis}: Finally, the tool output is injected back into the context. The agent synthesizes the final natural language response $y$, grounding its explanation in the generated trace $\tau$:
    \begin{equation}
        y = \mathcal{S}_{\theta_{small}}(Q, v, \tau)
    \end{equation}
\end{enumerate}

This design ensures that all business-critical logic remains within the verified, deterministic boundary of the DACL engine ($\Phi_{DACL}$), while the \texttt{gpt-5-mini} model handles the flexible user interaction at a fraction of the computational cost of the baseline models.

\section{Results and Evaluation}
\label{sec:results}

We evaluate our neuro-symbolic architecture against state-of-the-art frontier models on four proprietary industrial agreements. Our evaluation focuses on three key dimensions: (1) accuracy and consistency across test samples, (2) error taxonomy analysis, and (3) computational efficiency.

\subsection{Dataset and Experimental Setup}
\label{subsec:dataset}

The dataset consists of four real-world commercial agreements spanning healthcare, energy, and transportation. To preserve commercial confidentiality, we refer to them by their domain codes: \textbf{Health-PPO}, \textbf{Energy-Sup}, \textbf{Logistics-MSA}, and \textbf{Muni-IFB}. Table~\ref{tab:agreements} summarizes their domain, business criticality, and logical complexity.

\begin{table*}[t]
\centering
\small
\resizebox{\textwidth}{!}{
\begin{tabular}{|l|l|p{6cm}|p{6cm}|}
\hline
\textbf{ID} & \textbf{Domain} & \textbf{Business Criticality} & \textbf{Logic Complexity} \\ \hline
\textbf{Health-PPO} & Healthcare & Compliance \& liability; coverage determination and cost-sharing. & \textbf{High:} Nested boolean logic, multiple eligibility checks (age, relationship), and conditional copay waivers. \\ \hline
\textbf{Energy-Sup} & Energy & Procurement efficiency; calculation of price based on dynamic weight and market indices. & \textbf{Medium:} Arithmetic heavy formulas requiring precise unit conversions. \\ \hline
\textbf{Logistics-MSA} & Transport & Revenue assurance; calculation of complex transport rates and diesel surcharges. & \textbf{Very High:} Multi-branch decision trees (Day Rate vs. Mileage) with surcharge lookups. \\ \hline
\textbf{Muni-IFB} & Municipal & Contract management; automated invoice verification with time-varying fee schedules. & \textbf{Medium:} Date-based period lookups combined with index-based pricing formulas. \\ \hline
\end{tabular}
}
\caption{Industrial agreements used in evaluation. Complexity ranges from linear arithmetic to massive combinatorial decision trees.}
\label{tab:agreements}
\end{table*}

\paragraph{Contract Representation.}
We process raw OCR text for all agreements. The DACL pipeline translates these into a compact taxonomy of clause types: \textit{procedure} (sequential steps), \textit{logical\_clause} (branching if-then-else), and \textit{range\_clause} (continuous-to-discrete mapping). Appendix \ref{sec:dec_complexity} details the structural complexity of each agreement. Notably, \textbf{Logistics-MSA} presents a "Very High" difficulty profile with 76 distinct decision states (logic branches and pricing brackets).

\paragraph{Evaluation Pipeline.}
Our evaluation simulates a production environment using a three-stage pipeline:
\begin{enumerate}
    \item \textbf{Traffic Simulation}: We use a specialized Event
    Generator to create synthetic transaction data. The generator is
    seeded with scenarios drawn from live production traffic on the same four agreements, and parameters are varied within contract-valid ranges to produce 400 events (100 per agreement), covering every decision branch and pricing bracket.
    \item \textbf{Deterministic Ground Truth}: Ground truth is produced by a "Gold Standard" engine implementing the contract logic in deterministic code. The Gold Standard and the DACL graph were developed independently. Legal interpretation was provided by qualified lawyers and domain experts, and the resulting logic was then re-implemented independently of the DACL graph. This guarantees mathematical precision (e.g., strict rounding) and reproducibility.
    \item \textbf{LLM-as-Judge:} Following recent protocols for scalable evaluation \citep{zheng2024judging}, an independent model evaluates concordance between the system output and the Gold Standard.
\end{enumerate}

\subsection{Operational Accuracy and the Reasoning Cliff}
\label{subsec:operational_accuracy}

We evaluated system performance on a dataset of 400 randomized real-world events, simulating standard industrial traffic distributions. As shown in Table~\ref{tab:accuracy_comparison}, the \textbf{DACL Agent} achieved near-perfect reliability (99.5\% overall accuracy), effectively mitigating the stochastic risks inherent in pure LLM deployments.

\paragraph{The Complexity Invariance of DACL.}
Crucially, the DACL Agent demonstrated \textit{complexity invariance}: its performance remained stable ($>$98\%) regardless of whether the domain required simple arithmetic (\textit{Energy-Sup}, 1 state) or massive decision trees (\textit{Logistics-MSA}, 76 states). In contrast, frontier LRMs exhibited a sharp ``complexity cliff.'' While GPT-5.2 and Claude Sonnet 4.5 achieved parity with DACL on the arithmetic-focused \textit{Energy-Sup} ($>$99\%), they suffered catastrophic degradation on the \textit{Logistics-MSA}.

\paragraph{The Limits of Probabilistic Reasoning.}
The \textit{Logistics-MSA} domain serves as a stress test for the ``reasoning as compute'' hypothesis. Characterized by 76 distinct decision states and conditional surcharge tables, this domain overwhelmed the attention mechanisms of standard models. Notably, while the ``Medium'' reasoning effort improved GPT-5.2's performance on the temporal logic of \textit{Muni-IFB} dramatically (raising accuracy from 36\% to 95\%), it failed to rescue the model in the high-branching Logistics domain. This indicates that while Chain-of-Thought can effectively solve \textit{depth} (e.g., date lookups in Muni-IFB), it struggles with \textit{breadth} (e.g., maintaining state across the 28-branch decision tree of Logistics-MSA).

\begin{table}[h]
\centering
\small
% 1. Keep the vertical 
\renewcommand{\arraystretch}{1.3} 
% 2. Tighten horizontal space 
\setlength{\tabcolsep}{4pt} 
\resizebox{\columnwidth}{!}{
\begin{tabular}{l|cccc|c}
\toprule
\textbf{Agreement} & \textbf{GPT-5.2} & \textbf{GPT-5.2} & \textbf{Claude} & \textbf{Gemini} & \textbf{DACL} \\
\textit{(Domain)} & \textit{(None)} & \textit{(Med)} & \textit{Sonnet} & \textit{3 Pro} & \textbf{Agent} \\
\midrule
\textbf{Health-PPO} & 74\% & \underline{91\%} & 73\% & 69\% & \textbf{100\%} \\
\textbf{Energy-Sup} & \underline{100\%} & 99\% & \underline{100\%} & 91\% & \textbf{100\%} \\
\textbf{Logistics-MSA} & 22\% & \underline{46\%} & 45\% & 30\% & \textbf{98\%} \\
\textbf{Muni-IFB} & 36\% & 95\% & 93\% & \underline{96\%} & \textbf{100\%} \\
\midrule
\textbf{Overall} & 58.0\% & \underline{82.8\%} & 77.8\% & 71.5\% & \textbf{99.5\%} \\
\bottomrule
\end{tabular}
}
\caption{Accuracy on Test Events (n=400). \textbf{Bold} indicates best overall; \underline{underlining} indicates best baseline.}
\label{tab:accuracy_comparison}
\end{table}

\subsection{Anatomy of Logic Failures}
\label{subsec:failure_taxonomy}

To determine the root cause of the baseline failures, we employed a "Judge LLM" (GPT-5.1) to classify errors into three distinct operational categories: \textit{Variable Dependency} (VD), \textit{Distraction Hallucination} (DH), and \textit{Arithmetic Hallucination} (AH). Agreement-specific process guidance ensures the judge verifies not only outcome equivalence but adherence to the contractual calculation sequence.

\begin{figure}[t]
    \centering
    \includegraphics[width=0.79\linewidth]{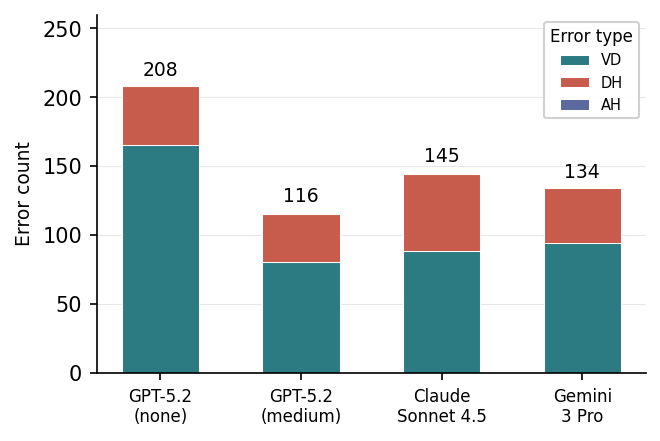}
    \caption{\textbf{Error Taxonomy by Model Configuration.} The stacked bars show the total failure count for each model ($N=400$ events). \textbf{Arithmetic (AH)} errors remain negligible across all models, confirming that the difficulty lies in state tracking rather than computation.}
    \label{fig:error_taxonomy}
\end{figure}

Figure \ref{fig:error_taxonomy} reveals a critical architectural insight: the primary failure mode of frontier models is \textbf{Variable Dependency (VD)}, which accounted for 71\% of all recorded errors across the four evaluated models. Conversely, \textbf{Arithmetic Hallucination (AH)} was statistically negligible ($<1$ error per model). This presents a paradoxical finding: frontier models have mastered the \textit{computational} primitives of law (arithmetic) but lack the \textit{structural} fidelity to apply them to the correct variables.

In complex environments like the \textit{Logistics-MSA} (characterized by 76 distinct decision states), models frequently calculated the math correctly but conditioned it on the wrong state - for instance, applying a "Tier A" rate to a "Tier B" shipment weight. This indicates that the difficulty of legal adjudication lies not in the complexity of the equations, but in the \textit{state tracking} required to navigate nested boolean logic.

\paragraph{DACL errors.}
The two DACL errors on Logistics-MSA both originated in the
orchestrator (\texttt{gpt-5-mini}), not in the symbolic engine, which
is deterministic by construction. One was a semantic routing error
where the agent selected the clause for a query that should
have resolved against a different clause; the other was a
tool-calling error where the agent invoked clause evaluation tool with an incomplete fact dictionary.
Both classes are addressable with stricter tool-input schemas.

\subsection{Computational Economics and Latency Profile}
\label{subsec:economics_latency}

\begin{figure}[t]
    \centering
    \includegraphics[width=\columnwidth]{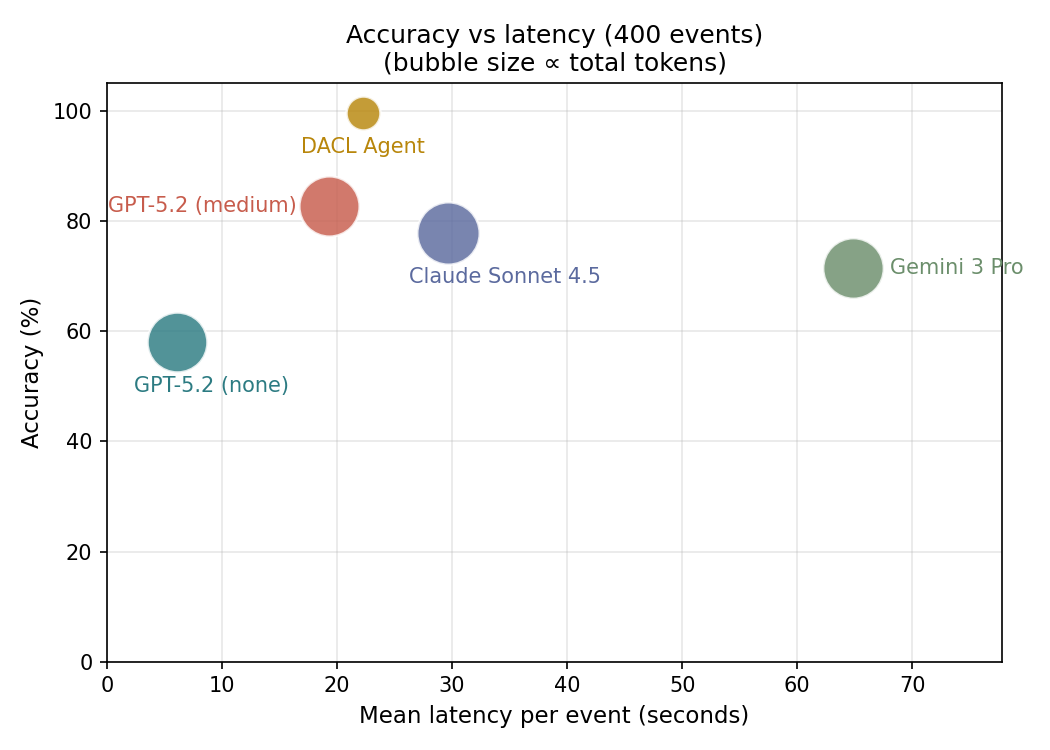}
    \caption{\textbf{Pareto Efficiency Analysis (Accuracy vs. Latency).} 
    Bubble size is proportional to total token consumption. The DACL Agent (gold) occupies the optimal quadrant, achieving near-perfect accuracy ($\approx$99.5\%) with moderate latency.}
    \label{fig:accuracy_latency}
\end{figure}

For industrial applications, the viability of an architecture is determined by its token economics and latency profile \citep{Chen2023FrugalGPTHT}. We analyzed the resource consumption for processing the full 400-event evaluation set.

\paragraph{Token Economy.}
The DACL architecture demonstrates a structural efficiency advantage. By retrieving only the active logic clauses rather than the full contract context, DACL reduced total token consumption by a factor of \textbf{9.9x} compared to the GPT-5.2 Medium baseline (1.35M vs. 13.44M tokens).

\paragraph{Latency vs. Reliability.} As illustrated in Figure~\ref{fig:accuracy_latency}, while the standard GPT-5.2 (No Reasoning) offered the lowest latency ($\sim$5--8s), its poor accuracy (58\%) renders it unsuitable for production. Crucially, the ability to offload complex reasoning to the deterministic DACL engine enabled the use of a lower-latency LLM (\texttt{gpt-5-mini}) for orchestration, avoiding the need for the heavy-duty reasoning models required by the baseline. Among the viable high-accuracy configurations, DACL provided superior throughput. On the complex \textit{Logistics-MSA} task, the DACL Agent averaged \textbf{26.8 seconds} per event, significantly faster than GPT-5.2 Medium, Claude Sonnet 4.5 and Gemini Pro ($\sim$164 s). 
This result suggests that shifting the "reasoning" burden from the LLM to a symbolic engine improves both speed and correctness simultaneously, avoiding the diminishing returns of extended inference observed in hierarchical legal reasoning \citep{zhang2025thinking}.

\section{Conclusion}

Current legal AI systems predominantly function as assistants, requiring a human expert to review and validate every output. While this ``human-in-the-loop'' paradigm is useful for individual practitioners, it does not offer a credible path to scalability or for tackling the systemic issues surrounding access to justice. Legal aid systems, courts, and public agencies must process millions of cases annually---a volume that renders manual review mathematically impossible.

To meet this industrial challenge, we cannot rely on probabilistic models alone. High-stakes adjudication requires a system that is fully automated, operates without human supervision, and remains consistent, explainable, and affordable at scale. Our work on \textit{Amortized Intelligence} demonstrates that achieving these requirements is feasible through a neuro-symbolic architecture. By treating the LLM as a one-time semantic compiler rather than a runtime interpreter, DACL eliminates the stochastic risks of ``System 2'' reasoning \cite{openai2025gpt52} reducing the marginal cost of adjudication to levels that make mass automation viable.

The implications of this architecture extend well beyond commercial contracts. The same deterministic approach demonstrated for Logistics and Healthcare agreements could be applied to other complex regulatory domains, such as civil procedure rules, tax code interpretation, city council bylaws, and social insurance claims. Extending the compiler target to defeasible and probabilistic
formalisms is a natural next step, for which DACL's structured audit trail provides the interpretability substrate.

Building such systems offers more than just efficiency; it offers stability, uniformity of rules, and speed in their application. Ultimately, by decoupling legal reasoning from the cost and unpredictability of generative inference, we pave the way for a legal infrastructure that is not only intelligent but structurally auditable and accessible to all.

\section*{Limitations}

While our neuro-symbolic architecture offers significant improvements in reliability and cost for industrial contract execution, several limitations remain.

First, the \textbf{expressivity of the intermediate representation} is currently bounded by the pre-defined primitives in DACL. Our system currently supports arithmetic, first-order logic, and range-based lookups, which cover the vast majority of commercial agreements. However, it does not yet support defeasible reasoning, open-textured legal standards (e.g., ``reasonable care''), or higher-order logic often found in statutory interpretation or case law. Expanding the primitive library to cover these broader legal domains remains future work.

Second, the \textbf{reliance on a ``Gold Standard'' for evaluation} restricts the scale of our testing. To rigorously measure correctness, we had to manually implement the ground truth logic for every agreement in code. This high annotation cost limits our ability to evaluate the system on thousands of different contract types simultaneously, a challenge common to neuro-symbolic research where automated ground truth is unavailable.

The traffic used in evaluation is synthetically generated rather than
sampled directly from production logs. Seeding from real scenarios
gives strong coverage of valid input variations and normal
operations, but it under-represents tail phenomena such as malformed
inputs or rare multi-amendment sequences that only a full
production-distribution study would surface.

Third, our experiments focused on \textbf{English-language commercial contracts} within specific domains (Logistics, Energy, Healthcare). We have not yet evaluated the performance of the semantic compiler on non-English jurisdictions or on informal, unstructured agreements that lack clear clause demarcations.

Finally, while the compilation step is a one-time cost, \textbf{compiler errors} can still occur. If the LLM misinterprets a clause during the initial translation to DACL, that error becomes deterministic and repeats for every subsequent transaction until corrected. Although our strict type system catches many such errors at compile time, human review of the generated graph is still recommended before deploying high-value contracts into production.
We describe the pipeline's qualitative error profile in
Appendix~\ref{app:pipeline-observations}.

Our LRM baselines use the full contract as context. A
retrieval-augmented variant that feeds the model only the clauses
relevant to a given event would be a stronger baseline. We expect it
to narrow the gap on depth-heavy clauses but not on breadth-heavy
ones, since Variable-Dependency errors arise from state-tracking
across dozens of branching nodes rather than from context length.
Quantifying this is left to future work.

\bibliography{custom}

\appendix

\section{LRM Experimental Settings}
\label{sec:models}

\begin{itemize}
    \item OpenAI GPT-5.2 (\texttt{gpt-5.2-2025-12-11}) \cite{openai2025gpt52}: Configured in two variants with default {reasoning\_effort="none"} and extended \texttt{reasoning\_effort="medium"} via the Responses API. This enables a balanced chain-of-thought depth suitable for industrial logic without excessive latency.
    \item Anthropic Claude 4.5 Sonnet \cite{anthropic2025sonnet45}: Configured with "Extended Thinking" enabled and a \texttt{max\_tokens=8192}. This allows the model to generate internal reasoning traces before emitting the final structured output.
    \item Google Gemini 3 Pro (\texttt{gemini-3-pro-preview}) \cite{google2025gemini3}: Configured with the default \texttt{thinking\_level="high"}, leveraging dynamic computation to traverse complex logical dependencies.
\end{itemize}

All baselines are constrained to output a strict schema containing a \texttt{reasoning} field (for auditability) and a \texttt{result} field (for automated evaluation).

\section{DACL Formal Specification}
\label{sec:appendix_dacl}

\subsection{Logical Clause Semantics}
Logical clauses represent branching conditions. Conditions are expressed in conjunctive normal form (CNF) with support for nested \texttt{AND}/\texttt{OR} blocks. The formal grammar for a condition $\phi$ is:

\begin{equation}
    \phi ::= \texttt{lhs} \bowtie \texttt{rhs} \mid \phi_1 \land \phi_2 \mid \phi_1 \lor \phi_2
\end{equation}

where $\bowtie \in \{=, \neq, <, \leq, >, \geq\}$.

\subsection{Range Clause Validation}
Range clauses map continuous inputs to discrete outputs. To ensure determinism, the DACL engine enforces the following constraints at load time:
\begin{itemize}
    \item \textbf{Non-Overlap}: No two brackets may cover the same numerical range.
    \item \textbf{Strict Inclusive Boundaries}: Intervals are evaluated as $\texttt{min} \leq x \leq \texttt{max}$.
    \item \textbf{Exhaustive Coverage}: Warnings are issued if gaps exist between brackets without a defined default fall-back value.
\end{itemize}

\subsection{Pricing Formula Sandbox}
Pricing formulas allow for arithmetic operations. To maintain security and determinism, the evaluation context is sandboxed. Only the following mathematical functions are whitelisted:
\begin{itemize}
    \item \texttt{ceil}, \texttt{floor}, \texttt{round}
    \item \texttt{sqrt}, \texttt{exp}, \texttt{log}
\end{itemize}
Arbitrary code execution (e.g., Python \texttt{eval}) is strictly blocked.

\subsubsection{Type System and Variable Model}

A DACL contract definition begins with a strongly-typed variable schema. Variables are categorized by \emph{source}:
\begin{itemize}
    \item \textbf{External}: Runtime inputs provided at evaluation time (e.g., shipment weight, service date)
    \item \textbf{Const}: Contract-specific constants with optional temporal validity windows
    \item \textbf{Derived}: Intermediate values computed during execution
\end{itemize}

 The engine performs runtime type coercion with strict validation. The validation layer intercepts malformed inputs in production deployments before they propagate to calculation steps.

\subsubsection{Temporal Validity and Amendment Support}
\label{sec: validity}

DACL natively supports clause versioning through \texttt{validity\_start\_date} and \texttt{validity\_end\_date} annotations. When multiple clauses share an identifier, the engine selects the appropriate version based on an \texttt{evaluation\_date} context parameter. The system enforces:
\begin{itemize}
    \item Temporal adjacency: No gaps between validity periods for clauses with the same name
    \item Non-overlap: At most one clause may be active for any given date
    \item Open-ended restriction: Only the chronologically latest clause may omit an end date
\end{itemize}

This enables mid-contract rate amendments (common in logistics and energy agreements) without requiring full contract regeneration.

\subsubsection{Execution Model and Audit Trail}

Each evaluation produces a structured audit trail containing:
\begin{itemize}
    \item \textbf{Variable states}: All input values with type validation results
    \item \textbf{Decision points}: For range/logical clauses, the specific condition or bracket that matched
    \item \textbf{Formula breakdown}: For pricing formulas, each sub-calculation with intermediate values
    \item \textbf{Execution path}: The sequence of clause identifiers traversed
\end{itemize}

\subsubsection{Error Handling and Recovery}

Rather than failing silently, DACL generates actionable error messages that guide correction:
\begin{itemize}
    \item Missing variables: ``Variable \texttt{X} required for clause \texttt{Y}. Expected: \texttt{type} - \texttt{description}.''
    \item Type mismatches: ``Variable \texttt{X} must be a number, got string \texttt{'abc'}.''
    \item Enum violations: ``Variable \texttt{X} has invalid value \texttt{'foo'}. Valid values: \texttt{A}, \texttt{B}, \texttt{C}.''
\end{itemize}

These messages are designed for programmatic consumption, enabling upstream systems (including LLM agents) to self-correct without human intervention.

\section{Decision state complexity}
\label{sec:dec_complexity}

Table \ref{tab:complexity} specifies the difficulty level for each benchmark scenario based on the underlying logic required to resolve the clause. We quantify complexity primarily through the number of ``States,'' which represents the count of distinct decision paths, condition combinations, or pricing tiers inherent in the source text. As the data indicates, the selected cases cover a broad spectrum of structural complexity, ranging from simple linear procedures (e.g., Energy-Sup) to high-dimensional logical ranges (e.g., Logistics-MSA with 76 distinct states).

\begin{table}[t]
\centering
\resizebox{\columnwidth}{!}{
\begin{tabular}{|l|l|r|r|l|}
\hline
\textbf{ID} & \textbf{Primary Type} & \textbf{Clauses} & \textbf{States} & \textbf{Output} \\ \hline
Energy-Sup & procedure & 1 & 1 & numeric \\ \hline
Logistics-MSA & logical + range & 2 & 76 & numeric \\ \hline
Muni-IFB & procedure & 1 & 3 & numeric \\ \hline
Health-PPO & logical & 1 & 5 & string \\ \hline
\end{tabular}
}
\caption{Decision state complexity. \textit{States} refers to the number of distinct logic branches (e.g., if/else paths) or pricing brackets}
\label{tab:complexity}
\end{table}

\section{DACL Compilation Pipeline}
\label{sec:dacl_pipeline}

The transformation of unstructured contract text into the deterministic DACL representation is achieved through a five-stage pipeline. This process ensures high fidelity between the source legal text and the resulting execution graph:

\begin{enumerate}
    \item \textbf{Agentic Contract Segmentation:} The raw document is first processed by a segmentation agent that decomposes the text into discrete, semantically coherent blocks, preserving hierarchical context (e.g., section headers and definitions).
    \item \textbf{Computability Classification:} Each segment is analyzed to determine if it contains computational logic (e.g., pricing formulas, boolean eligibility rules) or purely declarative prose. Only segments classified as \textit{Computational Legal Clauses} proceed to the translation phase.
    \item \textbf{Dynamic Few-Shot Generation:} We employ Anthropic Claude 4.5 Sonnet for the translation step. The model is prompted with domain-specific, dynamic few-shot examples—retrieved based on semantic similarity to the current clause—to generate the corresponding DACL primitive.
    \item \textbf{Automated Scenario Testing:} The generated DACL graph undergoes immediate verification. An auxiliary agent generates synthetic test vectors (edge-case inputs) to execute the graph, flagging runtime errors or type inconsistencies.
    \item \textbf{Expert Supervision:} Finally, the compiled logic is subject to human-in-the-loop review. Legal engineers validate the structural audit trail against the source text to certify production readiness.
\end{enumerate}

\section{Data: Sample Legal Clauses and Sample Events}
\label{sec:appendix_dacl}

\subsection{Sample Clauses Used}
Below we present an overview of the legal clauses found in contracts that we use for testing the DACL system. All clauses are taken from existing, real-world contracts that the DACL is processing in production deployment.

\subsubsection{Example Clause 1: Transportation Rate Clause in Logistics MSA}

This example illustrates a representative pricing clause commonly found in transportation and logistics master service agreements.

\begin{quote}
\textit{\textbf{3.8 Pricing.} All pricing -- including rates, charges (e.g., fuel surcharges), fees, or tariffs for the Vendor's services -- shall be as set forth in the Pricing Schedule and shall remain firm for the term of this Agreement, unless the Parties mutually agree in writing to amend it. If amended in writing, the rates and fuel surcharges in effect on the date the Bill of Lading is issued shall govern. Under no circumstances may the Vendor ``gross up'' the agreed fee for any taxes, fees, licenses, or other charges; the Client need not pay any such amounts and may apply any payments made toward future invoices as a credit. No additional rates, charges, fees, or tariffs shall be imposed by the Vendor without the Client's prior written authorization. Any rate or charge for oversized shipments or services not covered by the Pricing Schedule must be documented in writing (e.g., email) between the Parties, and a copy of that special rate agreement shall accompany the Vendor's invoice as supporting documentation.}
\end{quote}

Pricing is structured across multiple service categories:

\textit{Base Linehaul Rates}
\begin{itemize}
\item Tier A: Distance-based lift service for shipments up to 0.5 ton, charged at \$X.XX per mile plus fuel surcharge (Exhibit A).
\item Tier B: Distance-based lift service for shipments between 0.5 and 0.75 ton, charged at \$X.XX per mile plus fuel surcharge (Exhibit A).
\end{itemize}

\textit{Local Tariffs}
\begin{itemize}
\item Zone 1 (within 45 miles): Flat pickup and drop-off fees by equipment type (Exhibit B).
\item Zone 2 (45--140 miles): Flat local service fees by equipment type (Exhibit C).
\end{itemize}

\textit{Ancillary Fees}
\begin{itemize}
\item Backhaul discount: 50\% reduction of the applicable linehaul rate (Exhibit D).
\item Additional stops: \$100 per extra pickup or delivery location.
\end{itemize}

\textit{Fuel Adjustment}
\begin{itemize}
\item A fuel surcharge applies to all mileage-based charges.
\item The surcharge calculation method is defined in Exhibit E and locks at the Bill of Lading issuance date.
\end{itemize}

\subsubsection{Example Clause 2: Diesel Surcharge Estimation in Logistics MSA}

A diesel surcharge (or fuel surcharge) is an additional fee applied to transportation costs to offset fluctuations in diesel prices. The surcharge is calculated using published diesel price indexes and adjusted regularly, ensuring equitable compensation for fuel-related cost changes during transit.

\begin{quote}
\textit{The applicable percentage fuel surcharge (“FSC”) is tied to the U.S.\ DOE Gulf-Coast diesel index. Percentages are applied to all billable charges that consume fuel. Index values are reviewed weekly; the FSC in effect on the date of service applies.}
\end{quote}

Fuel surcharge percentages are determined by price brackets. For example:

\begin{itemize}
\item Diesel prices between \$2.00 and \$2.099 correspond to a 1\% surcharge.
\item Prices between \$4.10 and \$4.199 correspond to a 22\% surcharge.
\item Prices between \$4.50 and \$4.599 correspond to a 27\% surcharge.
\item Prices between \$6.60 and \$6.699 correspond to a 48\% surcharge.
\end{itemize}

Intermediate price ranges increase incrementally, typically by 1\% for each \$0.10 rise in fuel price.

\begin{quote}
\textit{Over \$6.70 per gallon, an additional 1\% surcharge applies for every \$0.10 increment.}
\end{quote}

\subsubsection{Example Clause 3: Natural Gas (LNG) Price Calculation in Energy-Sup}

Natural gas pricing relies on regional index values, conversion factors, and contract-specific adjustments derived from external data sources.

\begin{quote}
\textit{\textbf{Clause 3.3.1: Fuel Cost Determination}}

\vspace{0.5em}

\textit{All prices for fuel per unit shall be based solely on each month's respective regional gas index, with adjustments reflecting each month's index price.}
\end{quote}

\begin{quote}
\textit{\textbf{Clause 5.11: Pricing Methodology}}

\vspace{0.5em}

\textit{The price per unit payable by the purchaser is calculated as:}

\[
\text{Price per Unit} =
\frac{\text{Regional Gas Index}}{\text{Conversion Factor}} + Y + R + F
\]

\textit{where:}
\begin{itemize}
\item \textit{Regional Gas Index: average natural gas price for the delivery month.}
\item \textit{Conversion Factor: units per MMBtu of natural gas.}
\item \textit{$Y$: supplier costs and profit per delivered unit.}
\item \textit{$R$: fixed rebate per unit representing purchaser credit participation.}
\item \textit{$F$: freight cost per unit based on standard delivery volumes.}
\end{itemize}
\end{quote}

\begin{quote}
\textit{\textbf{Attachment A: Price Sheet}}

\vspace{0.5em}

\textit{Provides detailed values for variables $Y$, $R$, and $F$ across the contract term.}
\end{quote}

\subsection{Event Sample Data}

The events evaluated in this study are grouped into domains corresponding to the various contracts we process, e.g. : Logistics Master Service Agreements (Logistics MSA), Energy Supply Agreements (Energy-Sup), and Healthcare Preferred Provider Organization plans (Health PPO). Each event represents a concise factual scenario requiring adjudication against a corresponding legal clause. Conceptually, each evaluation asks:

\begin{quote}
\textit{Based on the provided event information and governing agreement, determine the applicable outcome for the specified clause.}
\end{quote}

For each contractual category, structured natural-language queries are generated to compute transportation rates, fuel surcharges, commodity pricing, or healthcare coverage determinations.

\subsubsection*{Logistics MSA Events}

Logistics events describe shipment-specific operational facts used to evaluate transportation rate clauses and diesel fuel surcharge provisions.

\begin{quote}
\textit{A 0.6-ton shipment travels 348 miles from Hawkins to a regional terminal and is delivered on January 17, triggering Tier B linehaul pricing and a diesel surcharge based on the weekly Gulf-Coast index.}
\end{quote}

\begin{quote}
\textit{A local pickup occurs within 40 miles of the origin facility with two additional delivery stops, requiring application of Zone 1 tariffs and ancillary stop fees.}
\end{quote}

\begin{quote}
\textit{A return shipment qualifies for backhaul discount following completion of a long-haul delivery under the base rate schedule.}
\end{quote}

\subsubsection*{Energy-Sup Events}

Energy supply events capture commodity delivery scenarios requiring application of regional gas index pricing formulas.

\begin{quote}
\textit{An LNG shipment is delivered to an industrial customer, with fuel pricing calculated using the monthly Southern California gas index and contractual conversion factors.}
\end{quote}

\begin{quote}
\textit{A bulk natural gas delivery incurs additional freight charges and supplier margin adjustments under the contract's pricing methodology.}
\end{quote}

\begin{quote}
\textit{A scheduled energy shipment applies a fixed rebate per unit based on Attachment A pricing tables for the current contract term.}
\end{quote}

\subsubsection*{Health PPO Events}

Healthcare events represent emergency medical scenarios requiring coverage determination under PPO plan provisions.

\begin{quote}
\textit{A 15-year-old child on a family plan is transported by ground ambulance for an emergency condition, receives medically necessary care in the emergency department, and is admitted to the hospital.}
\end{quote}

\begin{quote}
\textit{A 17-year-old dependent child on a family plan is airlifted to a hospital for emergency treatment and receives medically necessary services, despite the deductible not being met.}
\end{quote}

\begin{quote}
\textit{A 15-year-old child on a family plan is transported by ambulance but later determined to have received routine follow-up care rather than active emergency treatment.}
\end{quote}

Together, these scenarios provide the factual foundation required to evaluate contractual outcomes across logistics pricing, energy commodity calculation, and healthcare coverage. Distance values inform linehaul rates, fuel indexes determine surcharge applicability, regional gas benchmarks drive energy pricing, and patient and transport attributes govern healthcare eligibility. This structured representation enables consistent evaluation across domains, allowing models to apply contractual rules directly to operational and clinical facts.

\section{Compilation Pipeline: Qualitative Observations}
\label{app:pipeline-observations}

We report qualitative observations from operating the compilation
pipeline across production agreements. We frame these as a
limitation rather than a quantitative benchmark, since per-clause
compilation outcomes are not instrumented for clean error-rate
reporting.

\paragraph{Human review is mandatory.} Production readiness in this
setting does not require zero-error automation; it requires an
auditable artefact that a legal engineer can sign off on. Step 5 of
the pipeline enforces this review.

\paragraph{The type checker catches most structural errors.} Range
non-overlap violations, type mismatches in pricing formulas, and
missing \texttt{validity\_end\_date} on non-terminal clause versions
are caught at load time and surfaced to the compiler agent as
actionable messages, enabling
self-correction without human intervention.

\paragraph{Scenario testing catches most remaining semantic errors.}
Automated test-vector generation surfaces clauses whose runtime behaviour contradicts the synthetic cases derived from the clause text.

\end{document}